# Brain Inspired Face Recognition: A Computational Framework

Pinaki Roy Chowdhury, Angad Wadhwa and Nikhil Tyagi

**Abstract**

This paper presents a new proposal of an efficient computational model of face recognition which uses cues from the distributed face recognition mechanism of the brain, and by gathering engineering equivalent of these cues from existing literature. Three distinct and widely used features – Histogram of Oriented Gradients (HOG), Local Binary Patterns (LBP), and Principal components (PCs) extracted from target images are used in a manner which is simple, and yet effective. The HOG and LBP features further undergo principal component analysis for dimensionality reduction. Our model uses multi-layer perceptrons (MLP) to classify these three features and fuse them to form a sparsely connected model. A computational theory is first developed by using concepts from the information processing mechanism of the brain. Extensive experiments are carried out using eight publicly available face datasets to validate our proposed model's performance in recognizing faces with extreme variation of illumination, pose angle, expression, and background. We also investigated the same mechanism because of reasons discussed later on object recognition tasks as well. Results obtained are extremely promising when compared with other face and object recognition algorithms including CNN and deep learning-based methods. This highlights that simple computational processes, if clubbed properly, can produce competing performance with best algorithms.

## 1. Introduction

Human brain's capabilities are remarkable in handling almost every task, and it is brilliant in the task of face and object recognition. Even with minimal information (low resolution images) and extreme variations in illumination and pose, we are almost perfect in recognizing familiar faces. The challenge for building a perfect visual recognition system, which could rival or even surpass human brain's abilities, has been an active area of research for those working in the area of pattern recognition and computer vision. This is due to the inherent non-intrusive nature of facial recognition and its widespread applications in the area of biometric applications, law enforcement, deployment in real time applications (Zhao et al., 2003) etc. It forms a very crucial part of our social interaction, as in the case of faces it can reveal the identity and the emotional state of a person, like anger, fatigue, depression etc. There are basically two proponents of face perception mechanism in the brain, and they are either based on module or on specialized regions evolved for specific tasks (like face recognition) wherein brain acts as a general purpose computing machine whose areas are able to perform tasks across various cognitive domains (Kanwisher, 2006). The biological importance and applications have thus led to a huge surge in the development of fast automated recognition systems.

## 2. Related work

Studies (Turk and Pentland, 1991), (Bartlett et al., 2002), (Martinez and Kak, 2001) suggest that a large number of features do not necessarily increase the recognition rate of a classifier. Findings from human visual system (Sinha et al., 2006) help us to conclude a similar idea about curse of dimensionality. Sinha et al draws two important conclusions (Sinha et al., 2006) regarding the performance of human face recognition as a function of varying spatial resolution of images: a) Humans can recognize familiar faces even when images are of very low resolution, and b) High frequency information is insufficient for good face recognition. These results clearly indicate that human visual system relies on other cues or mechanisms to address the task of face recognition. Lee et al (Lee and Seung, 1999) introduced a biologically inspired dimensionality reduction technique which is able to learn about faces from its parts and not in holistic sense as in PCA or LDA. Non-negativity of their dimensionality reduction technique corresponds to the physiological fact regarding non-negative firing rate of neurons. A lower dimensional projection space effectively represents the different visual stimuli present in the brain (Lehky et al., 2014). Lehky et al provides the first estimate of dimensionality of object representation in the primate visual system keeping in account the increasing dimensionality with increase in sample size (Lehky et al., 2014). Bao et al suggests to map the lower dimensional object space which is supported by neuropsychological evidence (Bao et al., 2020). The concept is established and demonstrated to differentiate between animate, inanimate, stubby and spiky types of objects which have been observed to show similar neuronal activations (Bao et al., 2020). Different Principal Components (PCs) signify different dimensions across which the information is distributed and can be differentiated for objects as well (Bao et al., 2020).

Historically face recognition models or algorithms are designed either for better representation or classification using robust classifiers, and at times using both the above properties. Face recognition can be addressed using 2 methods: Image classification and image similarity. Our proposed framework works on the former and therefore we compare with similar experimental settings and datasets pertaining to image classification. Some of the recent methods work on image similarity and based on the current state of the art (SOTA), we compare with Arcface (Deng et al., 2019) by creating similar experimental conditions as explained in the results section below. We give a brief overview of such algorithms below. We would like to emphasize that some important and fundamental works in the area of face recognition is presented here and in no way, we claim that this is an exhaustive one. Readers interested in a rigorous review of face recognition algorithms are advised to look at these studies (Zhao et al., 2003), (Wang and Deng, 2021)

Rolls et al pioneered the representation of objects in terms of its dimensions (Rolls and Milward, 2000) (VisNet2) taking into account neurophysiological data from primate visual system. Visnet2 worked on the principle of dimensionality reduction of information going from retina to Inferotemporal Cortex (ITC) while extracting the required features thereby focusing on structural aspect of hierarchical model in the visual cortex.

Heisele et al used a SVM for multi-class face classification by adopting one vs. all strategy (Heisele et al., 2001). In this, *M* SVMs are trained where *M* is the total number of classes, and each of these *M* SVMs distinguishes a single class from the rest of the classes. A recent research breakthrough in development of face recognition classifiers came after development of Sparse Representation-based Classification (SRC) (Wright et al., 2009). SRC is an extremely robust classifier, capable of handling large illumination changes, extreme variations in expression and also large degree of corruption. In SRC, the query (testing) image to classifier is represented as a linear combination of all training sample used (from a face dataset). Serre et al proposed a biologically inspired method to define a hierarchical model of visual cortex (Serre et al., 2007). Task of object recognition is achieved by mapping the simple and complex cells through a combination of Gabor filters and pooling operations. Each layer is defined by different units which extract the features and are then classified using a support vector machine (SVM) which are then boosted to increase recognition accuracy. Déniz-Suárez et al used a nearest neighbor (NN) classifier (Déniz-Suárez et al., 2011) with Euclidean and cosine distance to classify faces by extracting HOG features (Dalal and Triggs, 2005) from facial images. Images were divided into small regions called cells with each cell having various patch sizes (in pixels). Final recognition rate is calculated by fusing the posterior probability matrix obtained from NN classifiers trained with different patch sizes using a product rule.

Most of the computationally (Heisele et al., 2001), (Wright et al., 2009), (Déniz-Suárez et al., 2011) and biologically (Rolls and Milward, 2000), (Serre et al., 2007) inspired recognition techniques discussed above perform well under controlled conditions and have their respective drawbacks. But none of them (except PCA, as will be discussed later) take cues from the face processing mechanism of our brain. Brain being the only organic machine which can handle the complexity of recognition almost perfectly, therefore it becomes a point in investigation to examine the efficacy of engineering systems that is constructed by taking cues - from functional perspective and not architectural one - from human brain. We, in this paper try to address the recognition problem, by introducing a new recognition mechanism called Brain Inspired Face Recognition System (BIFR). BIFR is a computational model of recognition, whose elements though not new, but are collated in a manner that is akin to the perception mechanism of our brain. According to Deng et al , “Face recognition is a complex pattern-recognition problem involved with early processing, perceptual coding, and cue-fusion mechanisms” (Deng et al., 2008), and therefore we examine the possibility of translating the aforesaid observation into an engineering design, in an approximate sense, and examine its performance computationally.

## 3. Engineering basis of biological process

Tsao et al suggested that humans and macaques share a similar brain architecture for the processing of visual stimuli (Tsao et al., 2003). Right at the inception we would like to draw the attention on a model of the distributed human neural system for face perception (Haxby et al., 2000). Here we have kept our modeling limited to the core system – termed as “Visual analysis (Haxby et al., 2000)”. This happens to be one of the fundamental works’ that is widely referred to and forms basis of our formulation. Though in another paper published later (Gobbini and Haxby, 2007) the same authors have slightly modified their core system, but essentially the focus remains to be “effective representational mechanism of individuals for analysis purpose”. With this idea in mind, we bring out the components of the core system for visual analysis by human in the following:

1. Lateral Fusiform Gyrus (LFG)
2. Superior Temporal Sulcus (STS)
3. Inferior Occipital Gyri (IOG)

We shall now examine the functional characteristics of the aforesaid components of the core system that is supposed to encode the “appearance of face”. To begin with we consider LFG. The face responsive region in the LFG is termed as Face Fusiform Area (FFA) and we focus our attention on nature of representation FFA extracts from faces and its primary properties. We do this to examine which mathematical representation(s) can actually emulate such properties with better approximation. It is reported by Kanwisher et al that FFA stores representation of faces in a manner that are partly invariant to simple image transformations such as changes in size, spatial scale and position but largely non-invariant to changes in most viewpoints and lighting direction of the face image (Kanwisher and Yovel, 2006),(Kanwisher et al., 1997). These facts are corroborated by Timothy et al (Andrews and Ewbank, 2004) whose FMR-adaptation experimental findings showed that FFA is not sensitive to image size and is sensitive to viewpoint and direction. FFA shows face-inversion effect (Haxby et al., 2000) (i.e. a higher response for upright than inverted faces) as well as holistic processing of faces. Isabel Gauthier et al (Gauthier et al., 1999), observed that in patients with prosopagnosia, object recognition impairment is significantly recorded with similar activations when compared to faces. This suggested that damage to FFA not only affected face recognition but visual recognition (including objects) as well. FFA (Gauthier et al., 2000) is also known to show activations for objects even though it has been well known for its role in facial recognition. Xu(Xu, 2005) further investigated to understand FFA’s role for non-facial images. Results of Xu suggests that there exists a significant activation in FFA region for non-face visual stimulus. Based on the aforementioned evidence, we propose to include an FFA equivalent in our model. Therefore, we need to identify those operators that vary with viewpoints and lighting direction. One operator that comes to mind immediately is LBP (Ojala et al., 1996),(Ahonen et al., 2006), which is not very robust against local changes in texture, caused for example by varying viewpoints or illumination directions. One also needs to figure out how LBP fares in terms of invariance to position, size and spatial scale. Position is not that significant as size and spatial scale, therefore we concentrate on the other two. In so far as scale is concerned it is reported by Mäenpää (Mäenpää and Pietikäinen, 2003) that large scale texture patterns can be detected using LBP, however, there are certain processing involved to it (Mäenpää and Pietikäinen, 2003). We will not go into that as it is not important for our research but wish to give a brief comment. If we examine the traditional LBP a bit closely, we will find that LBP features are composed as micro-patterns that are invariant to gray scale transformations. Therefore, the central issue in scale invariance remains that how many such micro-patterns are required to construct the full face? It is definitely not that one cannot construct large (scaled-up) or small (scaled-down) faces with respect to original faces by concatenating series of aforementioned micro-patterns. Based on above arguments we propose to model the structure of LFG (essentially, FFA) as LBP. Since LBP models texture, therefore we call this storage of pattern at LFG using LBP as “Spatial”.

Now we examine the role of STS from a similar viewpoint as that of LFG (FFA). STS is essentially responsible for storing of changeable aspects of faces that includes perception of eye gaze, lip movement and expression (Haxby et al., 2000). In FMR-adaptation experiments done by Timothy et al (Andrews and Ewbank, 2004), it is observed that the MR-activity of STS is similar to the brain region that processes changeable aspects of faces. It has been deliberated by Calder that FFA is responsible for coding facial identity whereas STS codes expressions (Calder and Young, 2005). But their report does not suggest any experimental evidence when coding of both identity and expression were attempted simultaneously. It is reported by Calder that studies have found that face responsive cells recorded in the STS were sensitive to various stimulus dimensions, for example, the global category – like human face, face of monkey, simple shape (Calder and Young, 2005) etc. It has also been observed that monkey identity, monkey expression, human identity, and human expression are also recorded in STS. Also, it has been observed that the poly-sensory properties of STS facilitate explanation of greater association of STS with facial expression and other changeable facial clues (Calder and Young, 2005). Therefore, it becomes necessary to include an engineering equivalent of STS’s functional behavior in our design. To decide engineering equivalent of STS amounts to identifying an operator that can extract orientation and direction of various micro patterns that forms in our face as and when we interact or communicate, or for that matter express ourselves on a certain issue. As STS is involved with information that looks for the directional nature of data – to translate a slew of such data items to information level where in direction-related properties are embedded within the data item – the first operator that comes in mind for capture of such type of data effectively is Histogram of oriented gradients. HOG has been previously used to effectively extract the changeable aspects of faces (Carcagnì et al., 2015) but has not been proposed as an engineering equivalent to STS. HOG first computes and then counts the edge orientations in small local patches that an image is normally divided for computation purpose (Dalal and Triggs, 2005). Essentially, in HOG one image is partitioned into many small patches wherein the HOG features are computed at each patch separately and finally they are combined to form the final HOG descriptor. HOG is an extensively tested and used operator; therefore, without delving in much detail about HOG we refer the interested readers to the paper by Pierluigi Carcagnì et al (Carcagnì et al., 2015) which reports a detail and elaborate study on applicability of HOG in facial expression recognition problem. Since HOG as a model of STS that essentially examines non-visual properties, therefore we call this storage of pattern at STS using HOG as "Numeric".

Lastly, we examine the role of IOG from a similar viewpoint as that of LFG (FFA) and STS but in brief. It is reported (Haxby et al., 2000) that IOG is responsible for early perception of facial features and the seminal and fundamental work (Turk and Pentland, 1991) deals with a computational approach for early and pre-attentive pattern recognition capability that is independent of 3D information or detailed geometry. Therefore, we feel that the functional characteristics of IOG can be aptly modeled using this approach (Turk and Pentland, 1991). This approach in literature is well known as eigenfaces which extracts principal components of original face image and uses the most informative ones for face coding and retrieval. IOG also plays an important role for structural similarity and spatial attention (Joseph and Gathers, 2003). It is reported by David Pitcher et al (Pitcher et al., 2011) that IOG either contains OFA (Occipital Face Area) or OFA lies in close vicinity of IOG. In literature OFA has been termed as functionally defined face selective region and it receives inputs from early visual cortex. It stores increasingly complex object shapes before these shapes are further analyzed in higher cortical regions. It is reported that the IOG region is especially sensitive to response towards eyes (Sato et al., 2016). As IOG is the most posterior brain region, which is sensitive to face related activations, many researchers share the opinion that IOG is involved in initial stage of face processing, particularly that of facial features – of which eye happens to be a very prominent one. A very interesting finding (Sato et al., 2016) states that IOG is not sensitive to eye gaze direction but eye in general. This establishes the need and usefulness of pSTS region as critical detail might be analyzed by combining features obtained from these two regions. IOG essentially enables rapid detection of eyes which subsequently enables the processing of eye direction. We shall call this third state as "Visual" due to obvious reasons mentioned by Turk (Turk and Pentland, 1991).

After having chosen descriptors that are aligned with biological processes, we examine the issue of large dimension of such data and propose to use PCA in both HOG and LBP feature space. The reason for choosing PCA is not arbitrary, but PCA is supposed to encode the perceptual front end. We shall discuss about this in the discussion Section. This will create most informative and a transformed feature space in lower dimension which will be tractable more easily. As deliberated by Turk (Turk and Pentland, 1991), that eigenvectors generated like that will be ordered and spanning the maximum possible information from information-theoretic viewpoint. Significantly, these transformed features may not represent our notion of a face per say or key components in a face descriptor. Nevertheless, they are able to capture the most relevant of them in an effective manner. We wish to examine efficacy of this in both LBP and HOG space. The motivation of this framework lies in studies of Haxby et al (Haxby et al., 2001) and others (Calder and Young, 2005),(Wallis, 2013) wherein it was discussed and deliberated at length about a unified framework for visual recognition and the role and significance of a framework like PCA in tasks of recognition. We therefore model the framework that is consistent with human visual analysis (core system) wherein models are first created using PCA, PCA-HOG and PCA-LBP; subsequently their outcomes are suitably fused to obtain the final result. We need to bring out here that there were earlier attempts in using PCA-LBP and PCA-HOG separately. Also, given the evidence found about these parts in the brain in their contribution in object recognition, we decided to carry out face-nonface and object recognition experiments as well. The reasons for choosing LBP and HOG along with PCA to form the core system responsible for visual analysis in the brain by giving suitable arguments, is the novelty of the work.

## 4. Methods

### 4.1. The Model Building Process

In all our face, face - nonface, object recognition experiments, we randomly partition the original input data into two parts: Training (50% of original input data) and the remaining for Testing. Training data which is used for model building is further partitioned into **90% for training/ learning the model** and remaining **10% for validation.** Test images are not used in the model building process, they are unfamiliar images fed to our proposed model which are used only to test recognition performance of our model, and we report the result on test set as our recognition rate. This two-step partition of input data is designed to suitably choose the feature space dimension to be used in our multilayer perceptrons.

Let's consider a simple example; if we have an original input image data of 200 samples, we first partition the data into 100 images (50%) for training and the rest 100 images for testing. From the 50% designated for training, we further partition it into 90% (X), i.e., 90 images for model building and the rest, i.e., 10 images for validating the built model. Performance on these validation images forms the stopping criteria for our network training algorithm(SCG) (Møller, 1993). When the validation error starts to rise, we stop the training of MLP and evaluate the MLP with their respective weights and bias configuration on the test set. However, it must be noted that for performance comparison of the proposed model with state of art algorithms recognition tasks, we use the exact number of training and testing samples as used by those authors with whom we compare our method (listed in results section and under the description of the dataset). To figure out the best configuration of our network, we generated large number folds (around 100) from the training dataset following the scheme presented in the aforementioned paragraph. We train the networks created by different number of hidden layer nodes, depending on the problem being handled, and report the best result out of those folds.

Our proposed computational model (BIFR) has six fundamental blocks:

## 4.2. Data Extraction

Vectorisation of images from the original datasets take place (2D image matrix to 1D vector) and these images are concatenated column-wise to form an input matrix. This input image matrix is partitioned further into two parts: training (training and validation) and testing.

## 4.3. Feature Extraction

Here, we start feature extraction process for our aforementioned three states – Visual, Spatial, and Numeric. Before features are extracted the input images are normalized based on the normalisation technique as discussed in Xudong Xie and Kin-ManLam (2006) (Xie and Lam, 2006), which we refer to as **LN** (Luminous Normalisation) during the task of face recognition. However, normalisation by standard method (zero mean and unit variance) is also performed on the images, referred as **SN** (Standard Normalisation) and the best results obtained, from either of the normalisation, are reported in Table 1. Due to paucity of space, LN cannot be discussed in details here; readers interested in its implementation are requested to refer to Xudong Xie and Kin-ManLam (2006) (Xie and Lam, 2006).

The approach discussed in this paper aims to capture the variations in the images, not only in pixel space but also in the relevant feature space and use this information to encode new (test) face images. To learn the three states, we train three separate MLPs for each of the three states using the SCG backpropagation algorithm (Møller, 1993) due to its better performance empirically, as it can be seen from Supplementary Fig. 1.

Principal Components are arranged in descending order of their importance (eigen value). After a certain number of PCs (which are data dependent) the individual contribution of a PC plummets and each subsequent PC gives diminishing returns. We select PCs to this point as selecting any more would increase the size of neural network without significant results. Since initial PCs encode maximum variance, for the case of MLP handling only PCA (PCA-MLP) while processing faces, we use the procedure described in Turk, M. & Pentland, A.(Turk and Pentland, 1991) to encode and compare our test images (See Fig. 1). We resize all our images in face datasets to 96×96 pixels for experimentation, except for datasets (MIT Faces and object datasets) which have their image size mentioned in their description. The image size 96×96 for face datasets was fixed for ease of calculation and to keep the parameters of the feature extractors standard. For performing LBP transformation on an image, we partition the image into 6×6 blocks (total 36 blocks), each block having a size of 16×16 pixels. LBP thresholding occurs in each of these blocks to extract the local histogram, which, are finally concatenated together to get the global histogram (See Fig. 2). One usually gets two type of patterns, while applying LBP thresholding operator in images: Uniform and Non-Uniform (Ahonen et al., 2006). In our proposed framework, we use uniform patterns which are assigned different histogram bins (label) and all other non-uniform patterns are assigned using a single label. Finally, we represent our LBP operator with the following notation, as described in the original text (Ahonen et al., 2006): $\mathrm{LBP}_{P,R}^{u2}$, where superscript $u2$ represent uniform patterns used and $P$, $R$ represent the number of sampling points chosen and size of radius respectively. To get the global histogram all local histograms are concatenated. We choose 8 sampling points ($P$) and circle radius of 1 ($R$) to calculate LBP feature vector in each of the 36 image blocks.

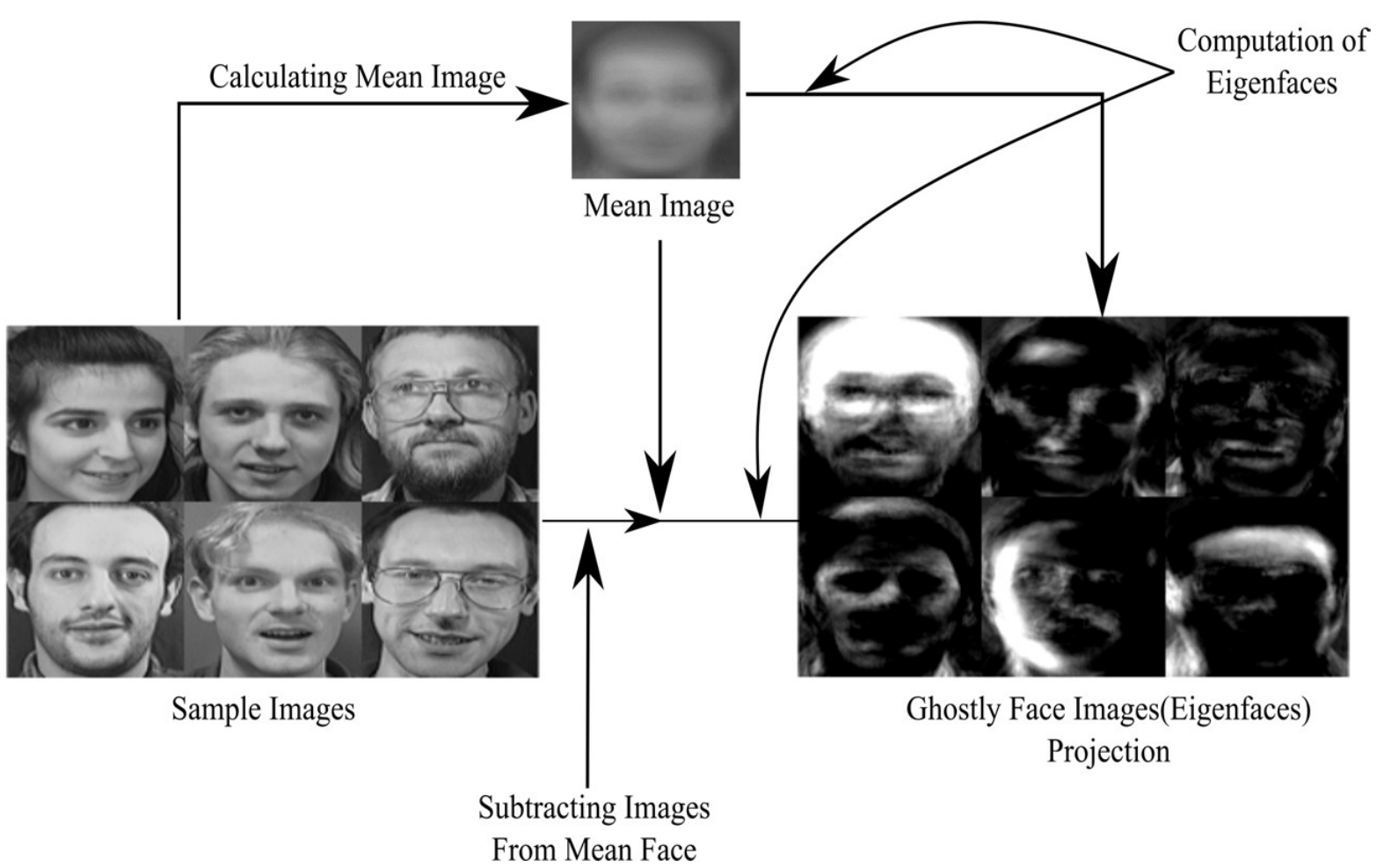

Fig. 1. Eigenfaces calculation on a sample input image of the AT&T dataset.

TABLE 1
TEST RESULTS FOR CORRECT CLASSIFICATION ON THE FACE DATASETS (IN PERCENTAGE)

| DATASET | RECOGNITION RATE | NORMALISATION | NO. OF PATTERNS USED FOR TESTING |
|---|---|---|---|
| AT&T | 98.50 | SN | 200 |
| EYB | 99.71 | LN | 1,207 |
| GT-C | 89.86 | SN | 375 |
| GT-F | 99.73 | NONE | 375 |
| MIT CBCL | 100 | SN, LN | 1,620 |
| CALTECH | 95.65 | LN | 161 |
| MIT FACES | 95.3 | SN | 5427 |

LFW has been excluded from Table 1 because – i) It's a matching problem and we are solving a problem of recognition. ii) On purpose due to multiple subsets chosen for the experiments performed.

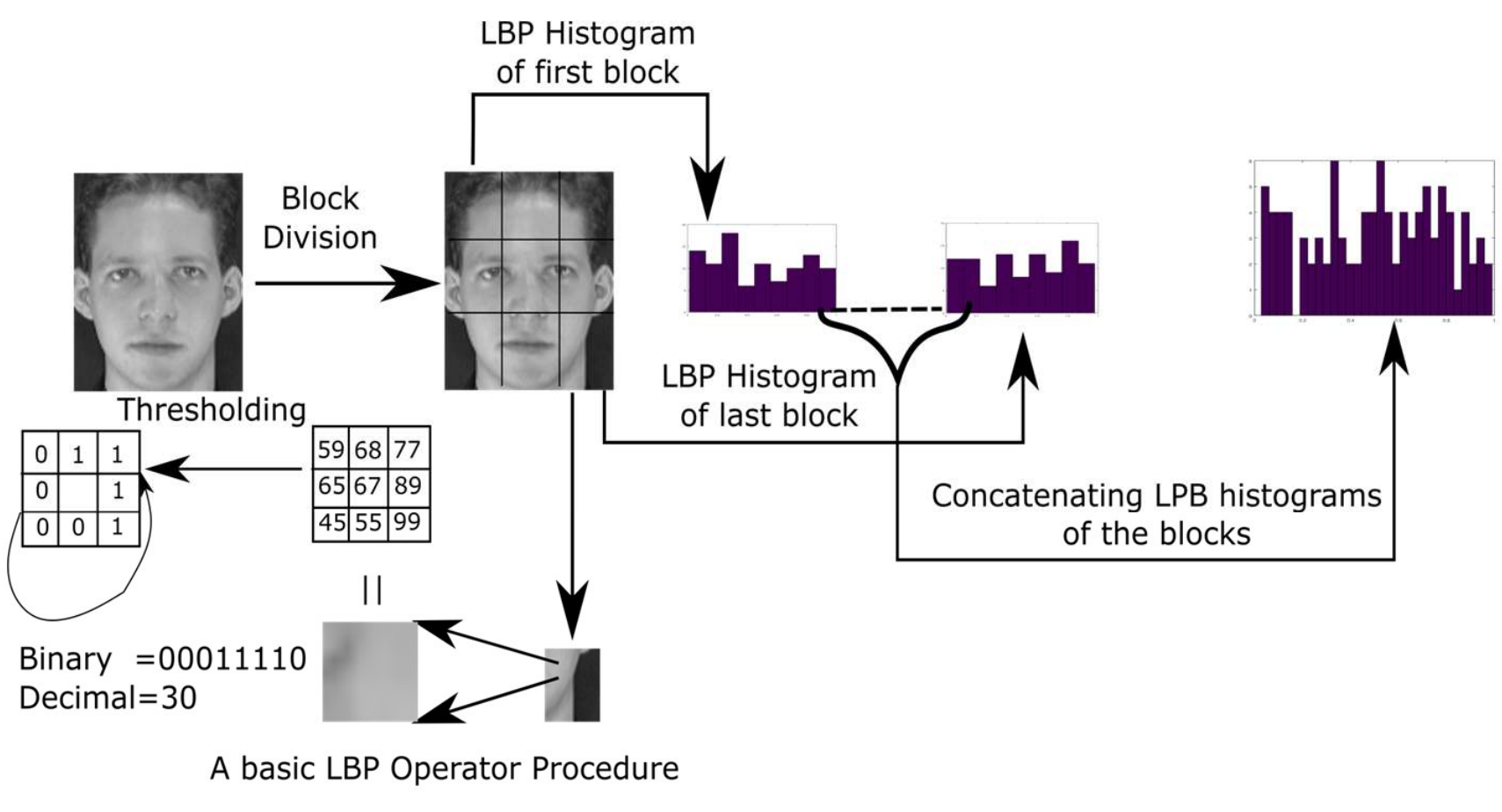

Fig. 2. LBP pattern histogram calculated on a sample input image of the AT&T dataset

In the case of HOG, an image is partitioned into smaller regions called cells. This paper uses cells arranged in a square of (8×8) pixels. A block of size 4, arranged as (2×2) cells, operated on these cells to calculate the HOG histogram. The histogram calculated has an overlap of 50% of cells from the previous block. Finally, all local histograms are concatenated to form a global feature vector (see Fig. 3). This is the most simplistic method and also the default method of calculating the HOG feature vector from the images (Dalal and Triggs, 2005). The number of bins used for the orientation histogram are 9 which quantize the gradients orientation in the range of 0 to 180 degrees.

### 4.4. Dimensionality Reduction

Images from feature extraction block (HOG and LBP) are subjected to PCA for dimensionality reduction. Each feature set is normalized to zero mean and unit variance. Eigen vectors are then calculated for normalized feature set via eigen value decomposition or Singular Value Decomposition. First X eigen vectors with largest eigen values are chosen by the procedure described in next sub section. Data is then transformed in terms of selected eigen vectors. PCA-MLP feature extraction and dimensionality reduction is as discussed above in the feature extraction section.

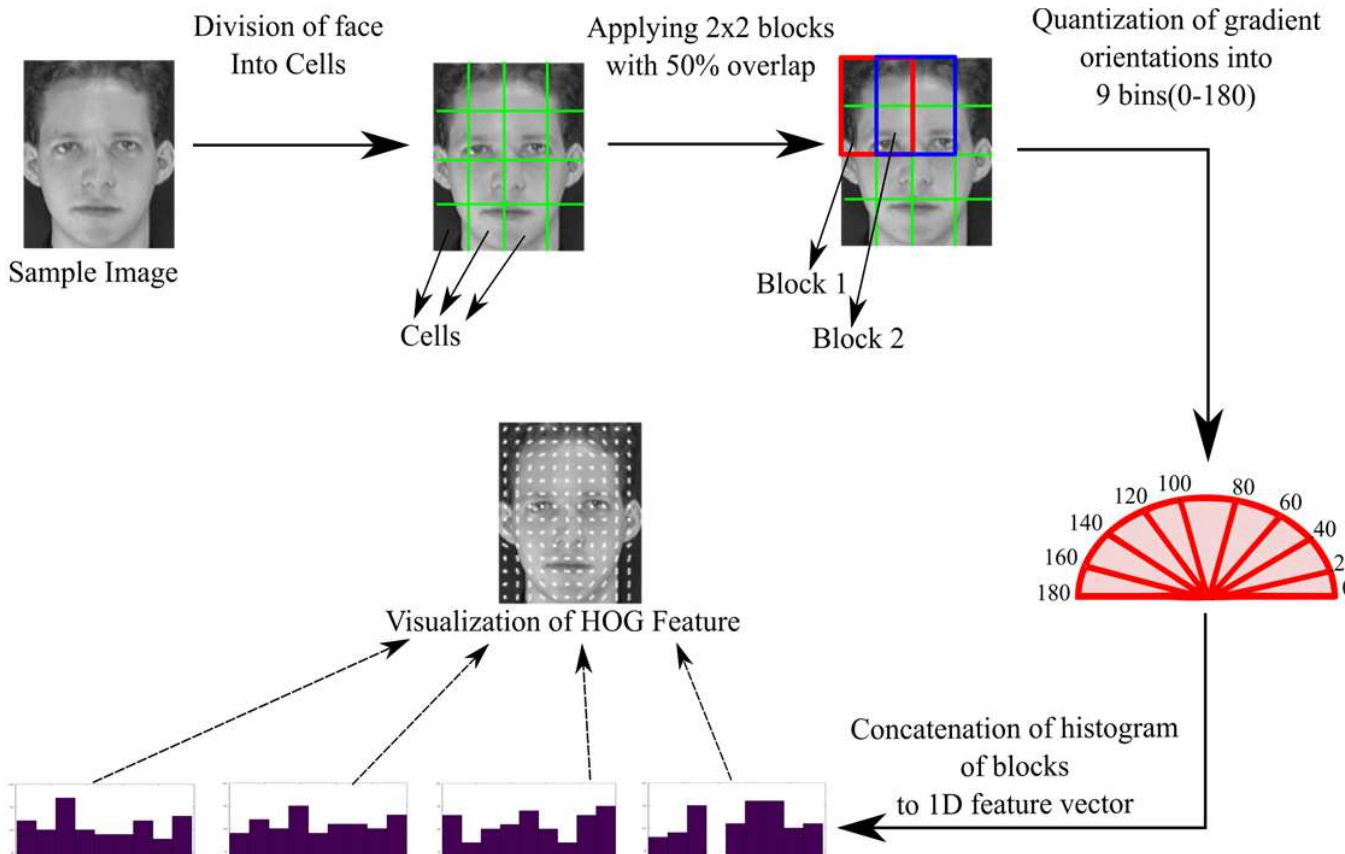

Fig. 3. HOG block operation applied on an image split into 6 cell regions. The block size chosen has 4 cells in it. This operation is performed on all the cells to get the global HOG histogram.

### 4.5. Architecture Selection

Each Network has two tuneable parameters, viz. number of PCs and number of neurons in the hidden layer. All possible combinations of PCs with neurons lying in range of 1-150 (PCs) and 1-100 (neurons) respectively are tested for all datasets using nested for-loops. This method can be extended for multiple trainable parameters, for example in case of two hidden layers (3 parameters), Grid Search (the mechanism discussed above) employs 3 nested loops. Architecture that performs best on validation dataset with least number of parameters is chosen.

### 4.6. Fusion

This strategy is inspired from the functioning of Committee Machines (Tresp, 2001) which is a powerful method to improve the final performance when using multiple classifiers. Committee Machines work on principle of different classifiers that can be aggregated to improve the performance of a model as compared to a single classifier. These classifiers function parallelly independent of each other.

Our strategy involves removal of output layers of individual MLPs and connecting hidden layers to a common output layer. This network Fused Hybrid Network (FHN) thus created has "*n*" set of inputs and one set of outputs, where "*n*" is the number of individual MLPs used to create the FHN. We employ two training strategies to train the FHN taking inspiration from deep neural networks (Roitberg et al., 2019), which are Fusion Pre-Trained (FPT) and Fusion Not Pre-Trained (FNPT). Fusion Pre-Trained refers to using weights of individual pre-trained MLP's as initial weights of fused network. Fusion Not Pre-Trained refers to random initialization of weights in the fused architecture. We created FHN with PCA-MLP, LBP-MLP and HOG-MLP. The fused architecture can be visualized from Fig. 5 where M1, M2, M3 are the different multilayered perceptrons' with P1, P2, P3 as inputs from the different feature extractors. The decision layer is cut and the hidden layers Q1, Q2, Q3 are then densely connected to output *R* respectively. The diagram clearly depicts input layer (P1, P2, P3) is sparsely connected to the concatenated hidden layer (Q1+Q2+Q3) which is connected to output layer having *R* nodes. This network has different components that interact with each other at the dense layer and not only at the decision level. This type of fusion relies more on the collective learning and distribution of weights as the weights of all the 3 components are fused in the hidden layer. This resembles more to the human brain given that all areas of the brain actively take part, and the weights can decide the importance each area gains through the process of recognition. We propose the sparsely connected MLP based model in this paper (Fig. 5).

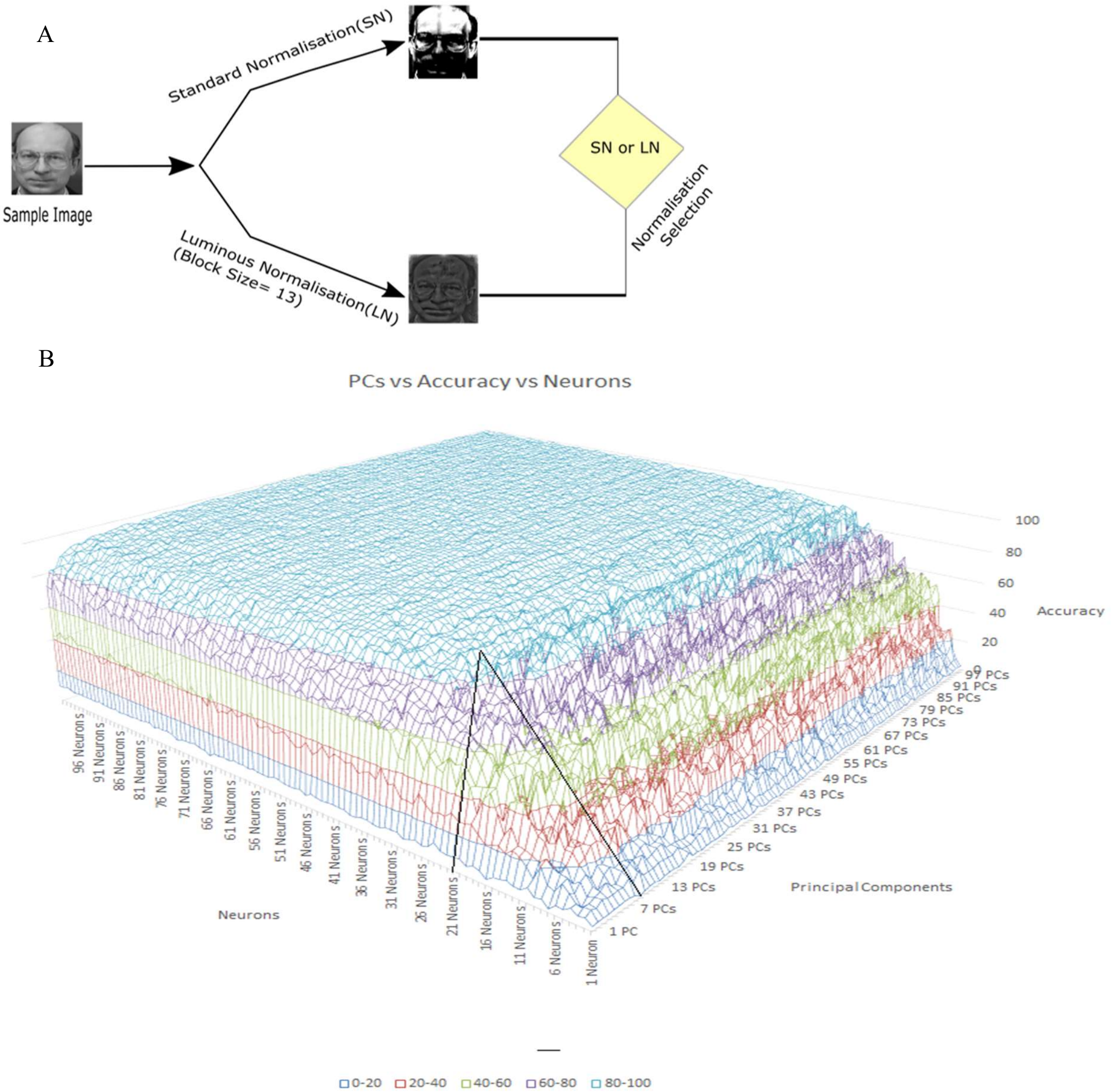

Fig. 4. (A) Luminous Normalization: Choosing SN or LN for pre-processing is purely on experimental basis and subjected to the dataset under consideration. (B) This graph is a plot for multiclass classification task – GT-F using 3 operator FPT. The X-axis represents the number of neurons, Y-axis the accuracy and Z-axis the number of principal components (PCs). The various colors signify the range of accuracy. The peaks and troughs indicate individual accuracies. The graph is a result of the grid search technique which is used to find the optimum set of parameters. The technique involves a unit increment across both (neurons, PCs) dimensions while comparing the accuracy for a given set of parameters. It can be observed that after 10 PCs and 21 neurons there seems to be no significant increase in the accuracy as compared to the changes observed in the first 10 PCs and 21 neurons. Light blue color on the peaks signifies the accuracy lies in the range of 80-100%.

## 4.7. Testing

After validation, which is performed as described in model building process, we test our model on the same testing protocols as described by the authors we compare with for fair comparison. A typical workflow of BIFR is given in Fig. 5.

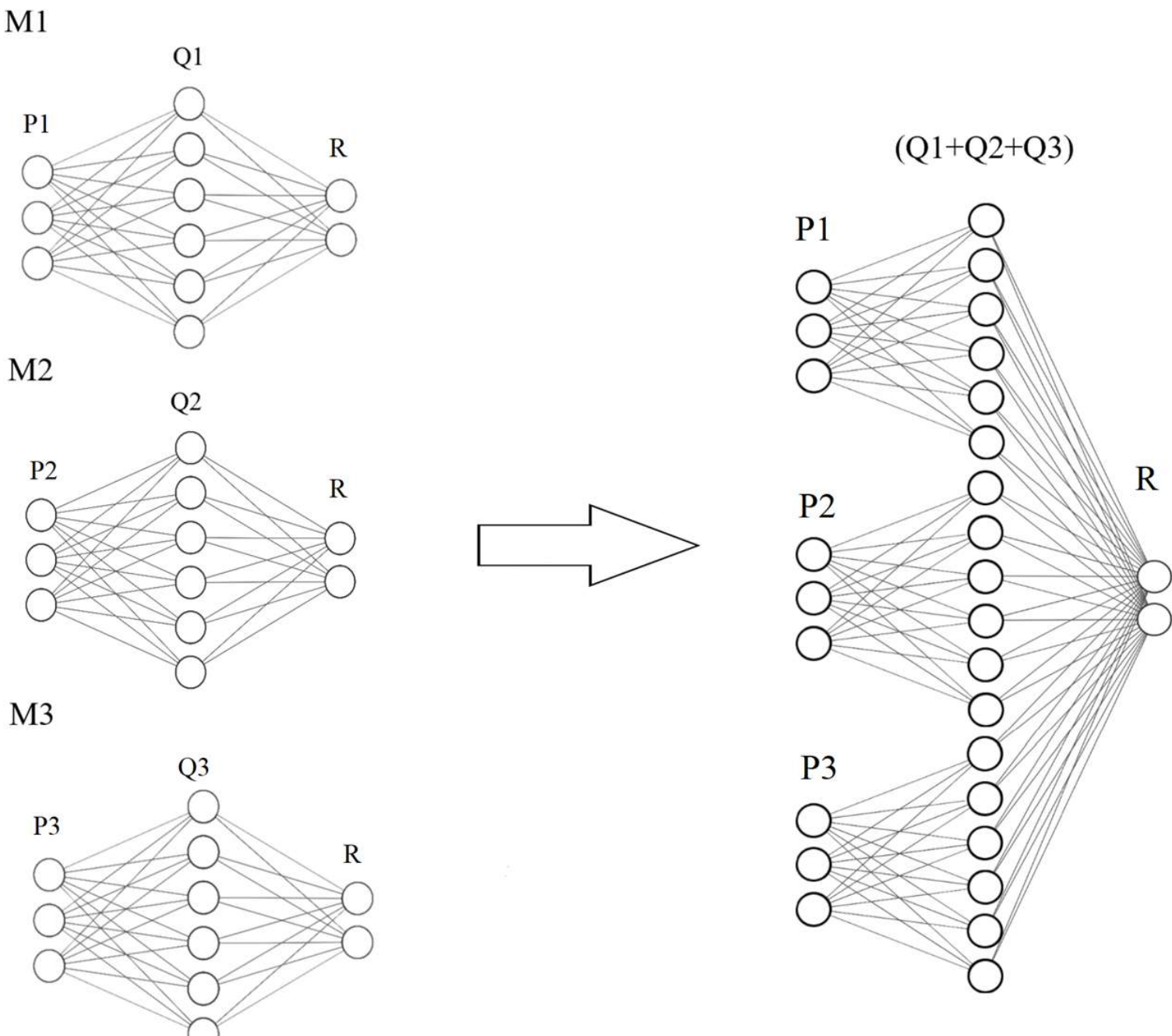

Fig. 5. Our proposed Fusion Pre-Trained (FPT) and Fusion Not Pre-Trained (FNPT) models have the same architecture and differ in the initialization of weights. Three different MLP's M1, M2 and M3 are fused. The decision layer (R) is cut while hidden layers (Q1, Q2, Q3) are fused and the input layer (P1, P2, P3) which are the outputs from different feature extractors is connected sparsely to the hidden layer and a new decision layer is formed. The weights from M1, M2 and M3 are used to initialize weights of the fused networks for FPT while for FNPT random initialization was done.

## 5. Results

### 5.1. Datasets

We use eight publicly available face datasets to compare and evaluate our proposed model with state-of-the-art face recognition algorithms. We also use face – non face and object datasets to evaluate our proposed model for reasons mentioned in the Recognition Results section. During pre-processing stage all the images in the cropped and un-cropped version were converted to grayscale. The face recognition tasks are normalised with SN or LN while the rest were normalized to zero mean and unit variance. Every face dataset belongs to face recognition category with MIT Faces being the only exception as it is a Face – Non-Face binary classification problem.

The Caltech faces (Weber, 1999) dataset consists of 450 images of 27 unique individuals. We randomly chose 17 images per class for those classes which have at least 19 images in them. The images are chosen randomly from each and every individual. There was a total of 19 classes or individuals who could satisfy this criterion. The dataset has extreme variations in illumination and background, pose and expressions. During pre-processing stage all images from this dataset were cropped (see Fig. 6) using the Viola Jones algorithm (Viola and Jones, 2001).

The Georgia Tech (GT) dataset ("Georgia Tech Face database," 2016) consists of 750 images (15 images per individual) of 50 individuals. The dataset has extreme variations in pose, expressions, and background. We conduct our experiments with both un-cropped (GT-F) and cropped version (GT-C) of the GT dataset. The cropped images are normalised with SN and the un-cropped images were used without any normalisation.

The AT&T Dataset (Formerly ORL Dataset) (Samaria and Harter, 1994) which is maintained at the AT&T Laboratories, Cambridge University; consists of 10 images for each of the 40 distinct subjects. The images are taken at separate times, and also under various lighting conditions. It characterizes changes in facial expression and facial details, such as glasses and no glasses.

The MIT CBCL dataset (Weyrauch et al., 2004) consists of 10 individuals having 324 images each. It contains synthetic images which are obtained from the 3D head models from these individuals.

The Extended Yale B dataset (EYB) (Georghiades et al., 2001),(Lee et al., 2005) consists of 2,424 images of 38 individuals (excluding bad, corrupted ones). For fair comparison, we dropped the 10 images which were completely black and randomly selected from the remaining 2,414 images in the cropped dataset and then performed all our experiments on them.

Labelled Faces in The Wild (LFW) (Huang et al., 2008) consists of more than 13000 face images collected from the web. The database was created to study the problem of unconstrained face recognition. Since the dataset was designed for face matching problem, we perform experiments on a subset of the dataset following Zhengming Li et al (Li et al., 2017) and others (Dora et al., 2017) for fair comparison with image size described in aforementioned papers (Dora et al., 2017; Li et al., 2017).

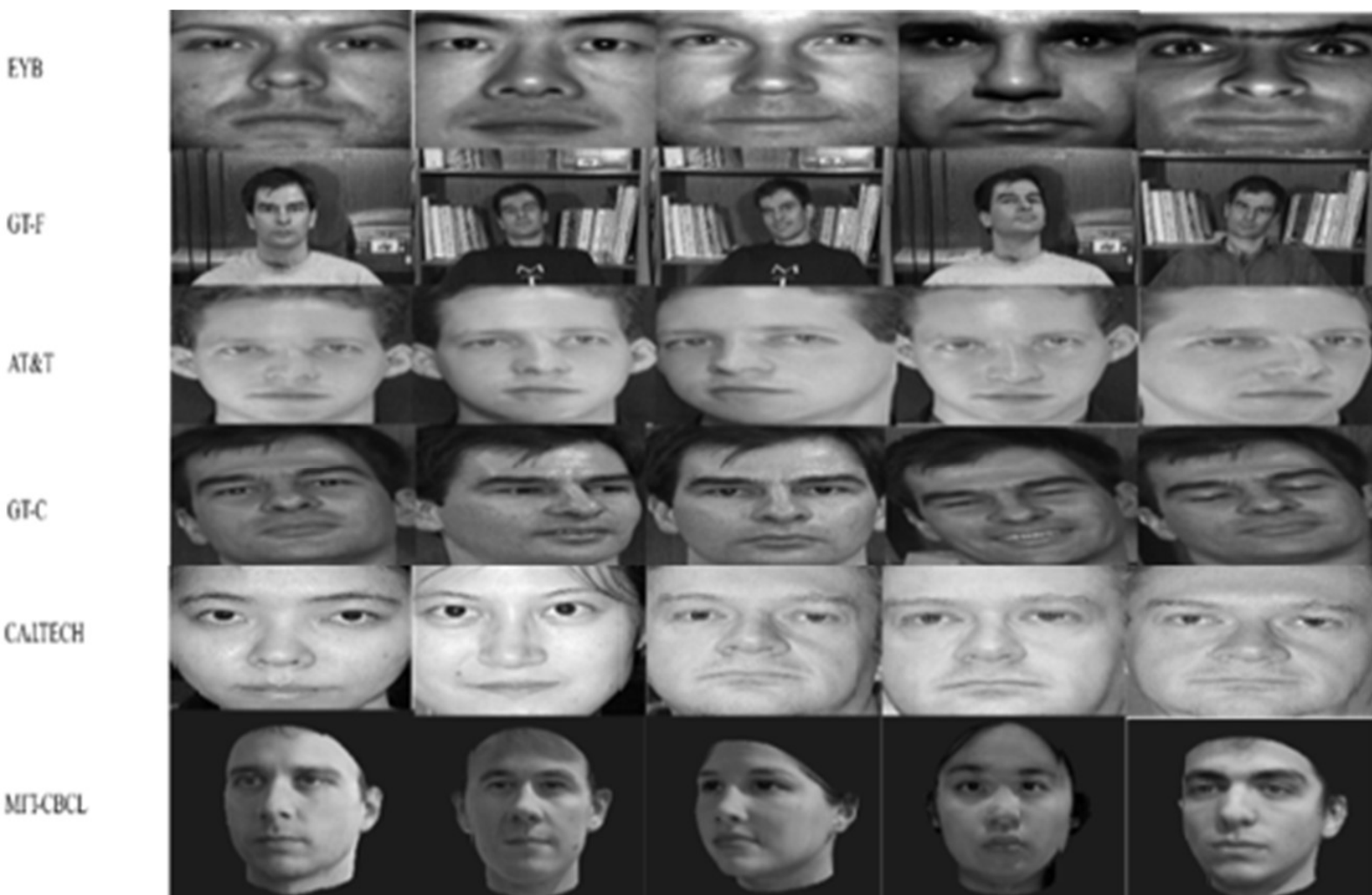

Fig. 6. Sample images of image datasets used in our experiments. The Caltech images shown here were cropped using the Viola Jones algorithm.

MIT Faces database has grayscale images of 2,429 faces and 4,548 non-faces for training and 472 faces and 23,573 non-faces for testing. Images in this dataset are rescaled to 48×48 pixels. Splits are taken as mentioned by Thomas Serre et al (Serre et al., 2007).

Caltech 5: Five Caltech datasets namely Airplanes, Motorbikes, Car-Rear, Faces and Leaves are taken from http://www.vision.caltech.edu/. Binary classification datasets are created for each category using the Caltech background dataset for negative samples. The binary classification splits are created as mentioned in Thomas Serre et al (Serre et al., 2007). Multiclass classification is performed by taking 15 samples per class for training and 50 (duplicates in case of less than 50 images) for testing. All Images in this dataset are rescaled to 192×192 pixels.

Caltech 101: Caltech 101 (Fei-Fei et al., 2004) has 101 object classes and one background class. Each class has a variable number of samples. Splits are created as mentioned by Thomas Serre et al (Serre et al., 2007) using background class for negative samples (binary) and using all 102 classes for multiclass classification. All Images in this dataset are rescaled to 192×192 pixels.

Caltech 9: This dataset contains classes of leopard, butterfly, chair, and sunflower from Caltech 101 alongside Caltech 5 dataset. 15 images from each class are chosen at random for training and 50 are chosen for testing (duplicates in case of less than 50 images). All Images in this dataset are rescaled to 192×192 pixels.

## 5.2. Recognition Results

On the Caltech dataset we randomly choose 17 images per class for those classes which have at least 19 images in them. The images were chosen randomly from each and every individual. There is total 19 classes or individuals who could satisfy this criterion. The size of the dataset used for comparison consists of 323 (19×17) images. The best recognition rate of 95.65% is obtained for 20 hidden nodes and with LN in the Caltech dataset. In MIT CBCL dataset, BIFR obtains perfect recognition rate of 100% for all the aforementioned hidden nodes sizes using both LN and SN. On the GT-F dataset, we obtained the best recognition rate of 99.73% for 64 hidden nodes and with LN. Experimentations on the GT-C dataset produced best recognition rate of 89.86% for 84 hidden nodes and with SN.

As can be observed from Table 1, in most of the datasets we achieved more than 95% successful recognition rate, GT-C being the only exception. These results establish the efficacy of our proposed model and its application in different types of facial images comprising large variations in illumination, pose, objects alien to faces such as glasses, etc. What is more interesting to note is how simple feature descriptors and elementary classifier(s) bundled together in a specific architecture, inspired by biological processes, can produce results that rivals the state of the art as discussed in the next sub-section. The approach adopted by us corroborates the fact that extraordinary face recognition results can be achieved using simple architectural configurations of multiple classifiers. These are shown as representative to assess the accuracy of our classification process.

We have performed extensive comparisons to test the performance of our proposed model on benchmark datasets. These comparisons are performed by creating a similar experimental set-up like those reported by the authors of corresponding algorithms with whom we report comparative performance. Emphasis is laid for maximum comparative evaluation with some recent state of the art algorithms. Since the comparison is quite exhaustive, therefore, it is difficult to adhere to all the pre-processing steps, resizing, cropping criteria, and dimension size as reported in the appropriate literature referred in this paper for comparison with BIFR. One important aspect to note is that there is no concept of validation set in the comparison experiments. We do not perform any comparison experiment on the Caltech dataset as we could not find any suitable algorithm or model which uses this dataset and report its result. We present our results in **bold face** and the best result in *italics* across all comparative studies. It is important to mention that due to paucity of space we are unable to include the full name of all the algorithms with which we compare our results. Due to the same reason, we include some of the detailed comparison tables for AT&T, GT-F, GT-C in the supplementary section. The images used per class from the datasets are indicated by x TRAIN in all the tables, for e.g., 4 TRAIN, 5 TRAIN etc.

The recognition accuracy on the GT-C dataset is compared with Tang et al (Tang et al., 2014) while that of GT-F dataset is compared with Naseem et al (Naseem et al., 2010) and Mohammed et al (Mohammed et al., 2011). The cropped images were normalised with SN and the un-cropped images were used without any normalisation. The neural network's (NN's) hidden nodes size is fixed at 30 hidden nodes for comparison experiments in both the cropped and un-cropped versions. We compare FPT, FNPT on different datasets in Fig. 8. We report the individual MLP accuracy as well to understand the contribution of each MLP and how the accuracy increases with fusion on LFW and GT-C dataset. The accuracy for the individual MLP's on LFW are as follows: 29.66% (PCA-MLP), 17.94% (LBP-MLP), 27.17% (HOG-MLP) while fusion achieves 48.13%. The accuracy for the individual MLP's on GT-C are as follows: 78.93% (PCA-MLP), 70.66% (LBP-MLP), 72.0% (HOG-MLP) while fusion achieves 89.86%.

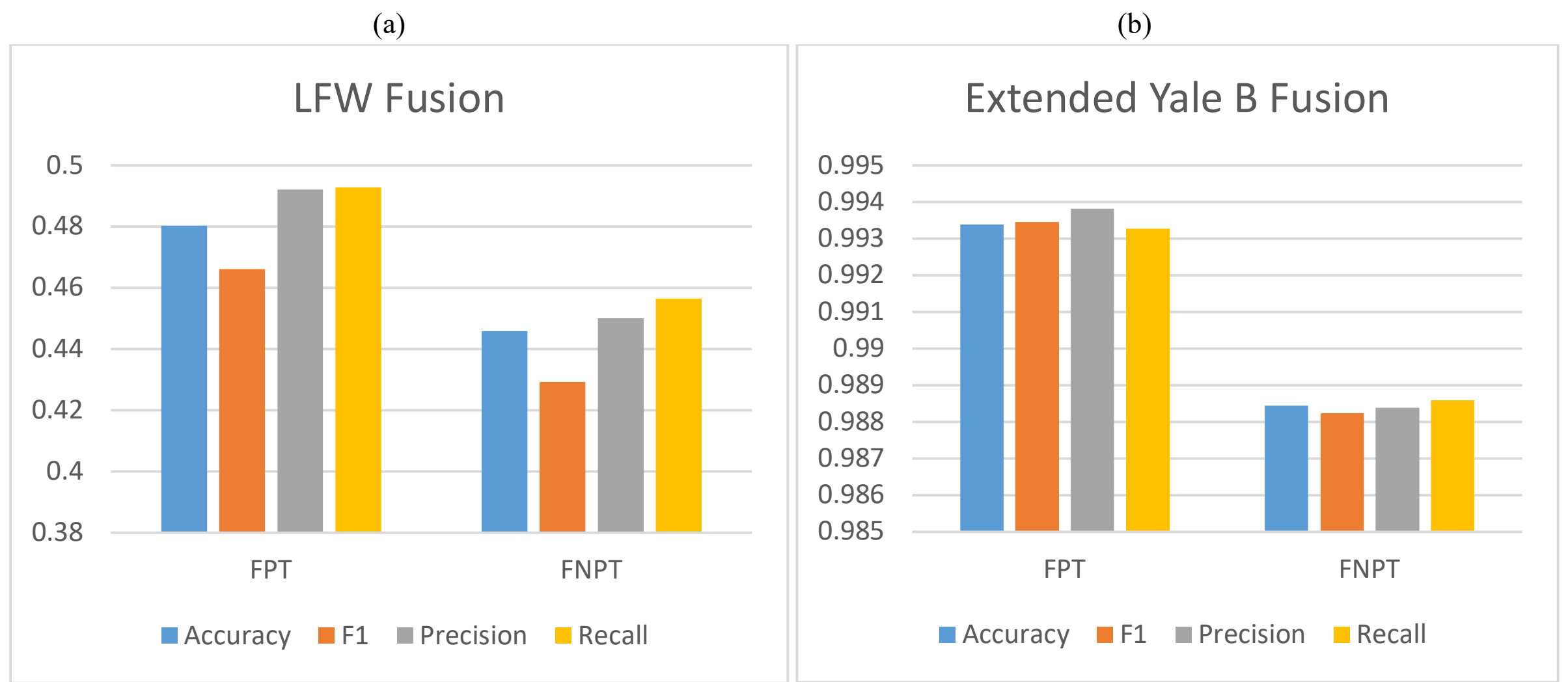

Fig. 8: Compares different Fusion techniques on LFW, Extended Yale B datasets in regard to Accuracy, F1, Precision and Recall. The results show how Fusion Pre-Trained (FPT) outperforms Fusion Not Pre-Trained (FNPT) on the aforementioned datasets. X-axis depicts different fusion techniques and Y axis depicts the accuracy in the range of 0-1. Different color-coding schemes depict different evaluation metrics.

We compare the result of BIFR on MIT CBCL dataset in Table 2. The comparison with the results reported by Zhang et al is accomplished by constructing the model by learning using 4, 5, and 6 samples per class only (Zhang et al., 2017). BIFR shows almost perfect recognition rate in all the three different number of training sample chosen per class. Nonetheless, we consider the MIT CBCL dataset an easy dataset as it does not contain extreme variations in pose and illumination conditions which are more likely to be encountered in real life situations. To understand the limitations of our model better, we chose 2 difficult datasets i.e., LFW and EYB.

TABLE 2
TESTING SET AVERAGE RECOGNITION RESULTS ON THE MIT CBCL DATASET

| ALGORITHM | 4 TRAIN | 5 TRAIN | 6 TRAIN |
|---|---|---|---|
| PCA(Turk and Pentland, 1991) | 0.6521±0.1128 | 0.7144±0.1379 | 0.7251±0.1415 |
| 2DPCA(Yang et al., 2004) | 0.7348±0.0031 | 0.8162±0.0040 | 0.8294±0.0056 |
| L1-2DPCA(Li et al., 2010) | 0.7049±0.0814 | 0.7912±0.0923 | 0.8176±0.0893 |
| S-2DPCA(Gu et al., 2012) | 0.6960±0.0868 | 0.8050±0.0939 | 0.8214±0.0951 |
| N-2DPCA(Zhang et al., 2015) | 0.7374±0.0037 | 0.8207±0.0090 | 0.8336±0.0056 |
| 2DLPP(Chen et al., 2007) | 0.8001±0.0968 | 0.8481±0.1045 | 0.8657±0.0953 |
| 2DNPP(Zhang et al., 2012) | 0.8284±0.0827 | 0.8789±0.0914 | 0.8816±0.0786 |
| 2DOLPP(Cai et al., 2006)(Kokiopoulou and Saad, 2007) | 0.6814±0.0800 | 0.7557±0.0944 | 0.8030±0.0901 |
| 2DONPP(Kokiopoulou and Saad, 2007) | 0.6912±0.0760 | 0.7703±0.0937 | 0.7922±0.0953 |
| N-2DNPP(Zhang et al., 2017) | 0.8608±0.1024 | 0.9104±0.0736 | 0.9050±0.0675 |
| S-2DNPP(Zhang et al., 2017) | 0.8493±0.0764 | 0.9098±0.0667 | 0.9032±0.0756 |
| **BIFR** | ***0.9950±0.0045*** | ***0.9994±0.001555*** | ***1±0.000*** |

Table 3 reports the comparison of BIFR on the EYB dataset. In Dornaika, F. & Khoder, A (2020) (Dornaika and Khoder, 2020) and others, (Zhan et al., 2019), (Wen et al., 2018) 20 images are randomly picked up per class for training and the rest of the images are used for testing. In Xia Wu et al (Wu et al., 2017) and others, (Wang et al., 2019),(Zeng et al., 2020),(Li et al., 2019) half of the images per class are used for training and the rest for testing. The authors in Dornaika, F. & Khoder, A. (Dornaika and Khoder, 2020) and others (Wen et al., 2018) reported only the mean accuracy and did not report standard deviation. In GSA (Majumdar, 2018) the author reports the best recognition accuracy of 98.1% while the best accuracy of our model is 99.86%. In Table 3, the average recognition rate across 10 random splits of testing are reported and compared with BIFR. BIFR outperforms all the reported algorithms. From the above tables we conclude that BIFR is robust to extreme variations in illuminations.

TABLE 3
TESTING SET AVERAGE RECOGNITION RESULTS ON THE EYB DATASET (IN PERCENTAGE). FIGURES IN BRACKETS INDICATE THE NUMBER OF DIMENSIONS USED

| ALGORITHM | RECOGNITION ACCURACY (20 TRAIN) |
|---|---|
| LRC(760)(Li et al., 2017) | 92.40±0.008 |
| LLC(Wang et al., 2010) | 88.90±0.010 |
| SRC(760)(Li et al., 2017) | 95.3±0.005 |
| K-SVD(456)(Li et al., 2017) | 94.0±0.005 |
| K-SVD(760)(Li et al., 2017) | 95.30±0.016 |
| D-KSVD(456)(Li et al., 2017) | 94.30±0.005 |
| D-KSVD(760)(Li et al., 2017) | 83.0±0.026 |
| LC-KSVD2(608)(Li et al., 2017) | 92.9±0.008 |
| LC-KSVD2(760)(Li et al., 2017) | 92.7±0.008 |
| LCLE-DL(722)(Li et al., 2017) | 95.4±0.005 |
| LCLE-DL(760)(Li et al., 2017) | 95.8±0.005 |
| RDA_FSIS (Dornaika and Khoder, 2020) | 95.11 |
| ICS_DLSR (Wen et al., 2018) | 96.80 |
| GLRRDLR (Zhan et al., 2019) | 96.42 ± 0.67 |
| **BIFR** | ***99.65±0.02*** |

Fig. 9 shows the performance comparison of BIFR and deep learning methods. For LFW we compare our model with Zhengming Li et al (Li et al., 2017) and others (Dora et al., 2017). The total number images are 1251 for 86 people where each person has 11-20 images each (Li et al., 2017). For training 8 (first 5 + 3 random) images are used and the rest for testing. The average recognition rate is reported with standard deviation. Dora et al (Dora et al., 2017) , used the same subset of images and performed 2 experiments. The first experiment has first 8 images for training and rest for testing while the second experiment has first 7 images for training while rest for testing. We compare our model with the same experimental procedure mentioned in both papers for fair comparison. Table 4 and Table 5 compare our model with Zhengming Li et al (2017) (Li et al., 2017) and others (Wen et al., 2018),(Zhan et al., 2019),(Li et al., 2019),(Dora et al., 2017).

TABLE 4
AVERAGE RECOGNITION RATE ON LFW WITH STANDARD DEVIATION (IN PERCENTAGE) (8 TRAINING SAMPLES)

| ALGORITHM | AVERAGE RECOGNITION RATE |
|---|---|
| LRC (Li et al., 2017) | 37.1±0.014 |
| LLC(Li et al., 2017) | 34.8±0.011 |
| SRC(Li et al., 2017) | 38.1±0.011 |
| K-SVD(Li et al., 2017) | 32.4±0.024 |
| D-KSVD(Li et al., 2017) | 33.4±0.016 |
| LC-KSVD(Li et al., 2017) | 32.2±0.012 |
| LCLE-DL (Li et al., 2017) | 36.8±0.013 |
| LCLE-DL (Li et al., 2017) | 38.8±0.009 |
| ICS_DLSR (Wen et al., 2018) | 44.47 |
| GLRRDLR (Zhan et al., 2019) | 46.07 ± 0.77 |
| LR-ASDL (Li et al., 2019) | 41.5 ± 0.01 |
| **BIFR (FPT)** | ***48.02±2.103*** |

TABLE 5
RECOGNITION RATE ON LFW FOR EXPERIMENT 1 AND 2 COMPARED WITH DORA ET AL(Dora et al., 2017) (IN PERCENTAGE)

| ALGORITHM | EXPERIMENT NO 1 (8 TRAINING SAMPLES) | EXPERIMENT NO 2 (7 TRAINING SAMPLES) |
|---|---|---|
| PCA + MDC(Dora et al., 2017) | 15.99 | 15.87 |
| LDA + MDC(Dora et al., 2017) | 10.43 | 11.09 |
| GABOR FILTER BANK (3X5) + MDC(Dora et al., 2017) | 10.12 | 09.40 |
| GABOR FILTER BANK (5X8) + MDC(Dora et al., 2017) | 16.87 | 14.79 |
| AGGREGATED 2D GABOR FEATURE METHOD + L2 NORM(Dora et al., 2017) | 18.83 | 16.95 |
| ESGK (Dora et al., 2017) | 39.08 | 37.5 |
| **BIFR (FPT)** | ***47.42*** | ***46.22*** |

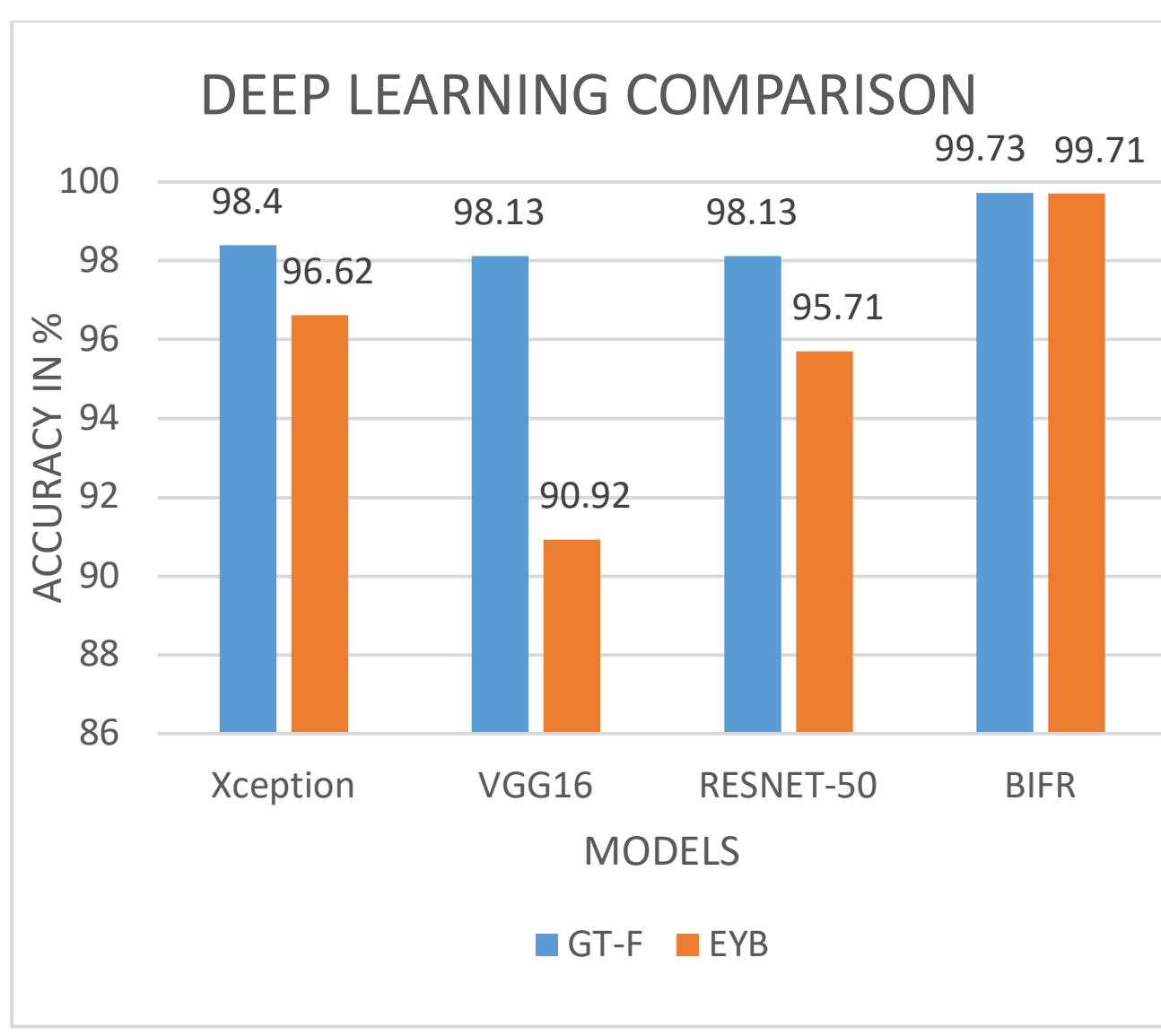

Fig. 9. Performance comparison of deep learning models (Resnet-50, VGG-16, Xception) and BIFR (FPT) on Georgia Tech Full (GT-F) and EYB.

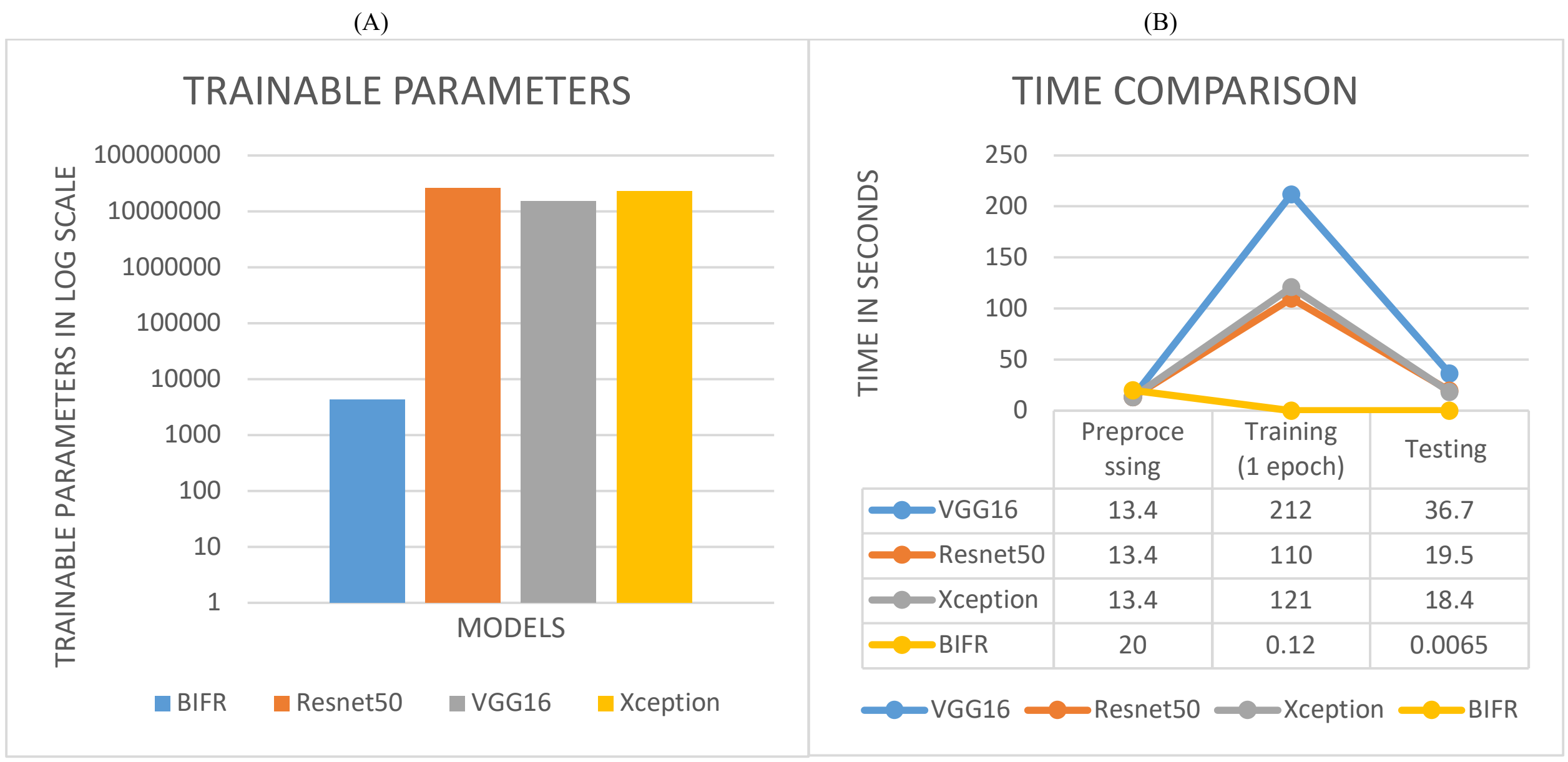

Fig. 10. (A) Comparison of number of trainable parameters of deep learning models vs BIFR (FPT) on logarithmic scale. (B) Comparison of pre-processing time, training time (1 epoch), testing time of our model (FPT) with deep learning methods in seconds.

Fig. 10 is a comparison of the number of trainable parameters of our model and deep learning models. We can infer that our model outperforms VGG16 (Simonyan and Zisserman, 2014), Resnet50 (He et al., 2016) and Xception (Chollet, 2017), while being orders of magnitude smaller in terms of model capacity. We also compare our model with ArcFace but since that model is not meant to solve face recognition problem but is rather developed to solve face matching problem, we used Support Vector Machine (SVM) as a classifier to classify the feature vector output given by ArcFace. Which we believe is not a fair comparison, ArcFace still gives a decent accuracy of 84.9% on EYB datas
et on which our model achieves accuracy of above 99%.

We report the best results obtained on the test set for the three object datasets (Caltech 5, Caltech 9, Caltech 101) in Table 6 below alongside the number of hidden neurons used. Our model is not an object recognition model, but as studies suggest that there is weak involvement of face recognition regions of the brain in object recognition task as well, therefore we investigate on object datasets to examine the efficacy of the aforesaid theory.

TABLE 6
TEST RESULTS FOR CORRECT CLASSIFICATION OF BIFR ON THE THREE CALTECH OBJECT DATASETS (IN PERCENTAGE)

| DATASET | TEST ACCURACY OF **BIFR** | NO OF TEST PATTERNS USED | NEURONS |
|---|---|---|---|
| CALTECH 5 | **98** | 250 | 29 |
| CALTECH 9 | **92.223** | 450 | 64 |
| MOTORBIKES | **99.37** | 726 | 61 |
| AIRPLANES | **99.35** | 974 | 44 |
| CARS | **100** | 426 | 44 |
| FACES | **100** | 350 | 44 |
| LEAVES | **99.68** | 86 | 44 |
| CALTECH 101 | **43.9** | 5050 | 505 |

In Table 7 we compare our model against Thomas Serre et al (Serre et al., 2007) and others (Holub et al., 2008), clearly showing our model outperforming others. Fig. 11 shows the performance comparison of BIFR and deep learning methods. Table 8 includes the results of our model in comparison with deep learning-based methods on object datasets.

Table 7
Comparison of binary classification accuracy (FPT) on Caltech 5 dataset with previously published methods. Test Set best recognition results using 100 samples per class for training and rest for testing (in percentage)

| Dataset | Holub (Holub et al., 2008) | Poggio (Serre et al., 2007) | BIFR |
|---|---|---|---|
| Cars | Not Available | 99.8 | ***100*** |
| Caltech Faces | 91 | 98.2 | ***99.71*** |
| Motorbikes | 95.1 | 98 | ***99.07*** |
| Airplanes | 93.8 | 96.7 | ***99.24*** |
| Leaves | Not Available | 97 | ***99.77*** |
| Mit Faces | Not Available | *95.9* | **95.3** |
| Caltech 101 | Not Available | *44.0* | **43.9** |
| Average all | Not Available | 89.94 | ***90.99*** |

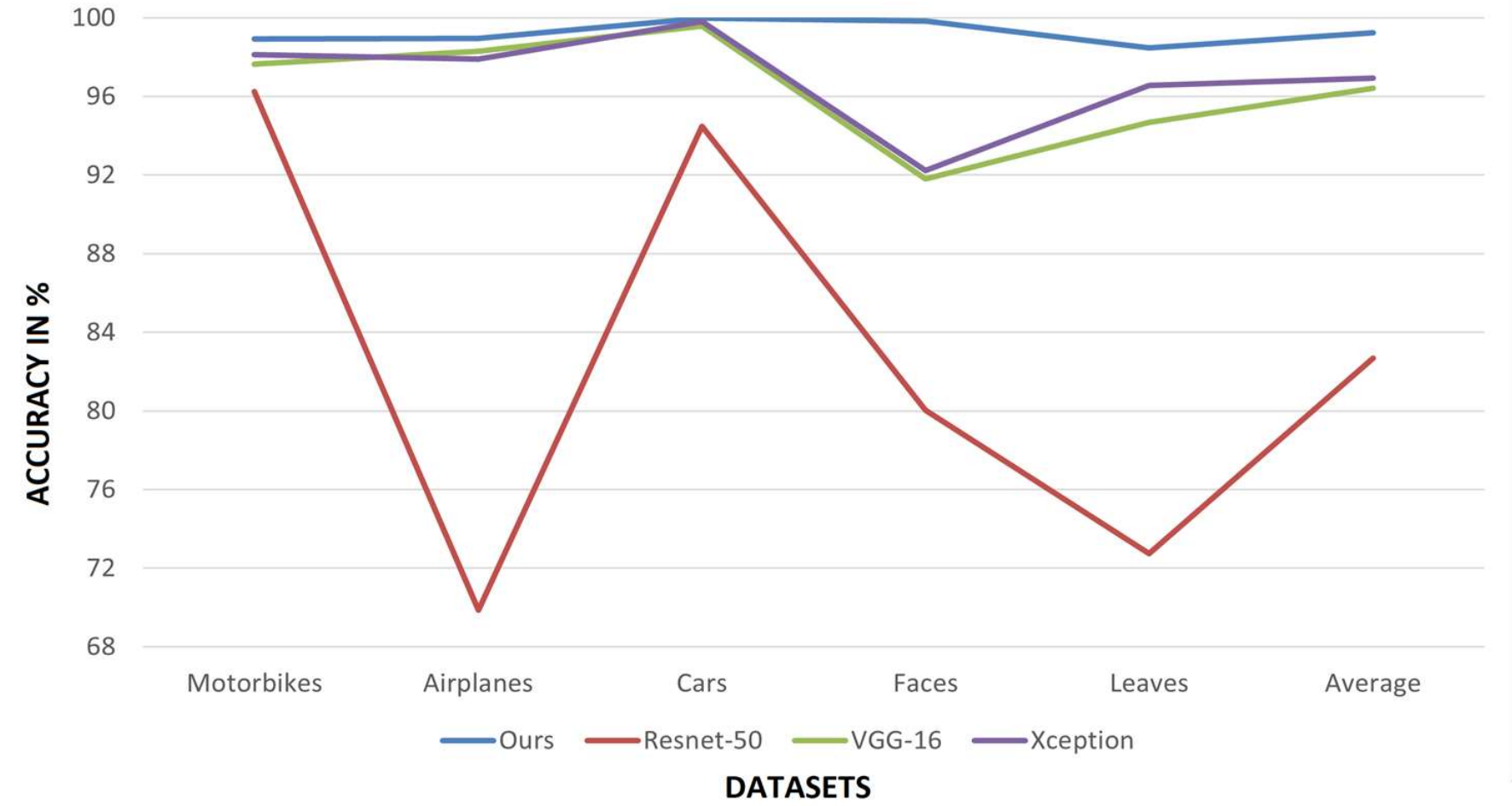

Fig. 11. Performance comparison of deep learning models (Resnet-50, VGG-16, Xception) and BIFR on Motorbikes, Airplanes, Cars, Faces, Leaves and average all these objects in binary classification. Our model outperforms other methods in terms of accuracy.

Table 8
Results of object categories in comparison with deep learning models (in percentage)

| Dataset | **BIFR** | Resnet-50(He et al., 2016) | VGG-16(Simonyan and Zisserman, 2014) | Xception(Chollet, 2017) |
|---|---|---|---|---|
| Motorbikes | ***98.918*** | 96.24 | 97.65 | 98.12 |
| Airplanes | ***98.937*** | 69.86 | 98.3 | 97.9 |
| Cars | ***99.979*** | 94.48 | 99.59 | 99.8 |
| Faces | ***99.824*** | 80.04 | 91.8 | 92.24 |
| Leaves | ***98.464*** | 72.73 | 94.67 | 96.55 |
| Average (of all above objects) | ***99.2244*** | ***82.67*** | ***96.402*** | ***96.922*** |

All Deep Learning models were imported from Keras and for computation purposes we used a Linux desktop with the following specifications: Intel i7 10700 Octa Core, 16GB RAM. Model Implementation is done in Python 3.

## 6. Discussion

In this work we examined engineering possibility of a biological reality. Our work presents a new direction in modeling computations of the Human Visual System (HVS), by examining the various functional aspect of areas of human brain. As reported by Contini et al (Contini et al., 2020), no single feature can represent the multidimensional visual representation in the brain. This study correlates with our proposal wherein multiple feature extraction algorithms are used which clearly removes the dependency on a single type of feature. Our aim is not only to provide mathematical operations with a classifier for the relevance of face recognition but to establish the underlying principles in the human visual system that correlates with contributions made in the field of neuroscience and establish a functional backbone that supports our knowledge and discoveries. Human eye captures incredible detail which far exceeds a digital camera (Skorka and Joseph, 2011). Details of an image are directly related to information, if deep neural networks are any indication, the requirement of neurons scales up with increase in information. Hence HVS must employ some form of dimensionality reduction before classification. Upon examination of host of literature, we noticed that there is a strong emphasis on the significance and role of PCA in our biological processing of faces per se. Notwithstanding conflicting but demonstrable evidence for views regarding the possibility of usage of a mechanism like PCA in our brain for face recognition, we decided to design a system whose foundations are laid down in Engineering Basis of Biological Process section. The authors in Andrew J. Calder & Andrew W. Young (2005) (Calder and Young, 2005) and others (Burton et al., 1999),(Bao et al., 2020) presented the importance of PCA at sufficient details in the domain of psychology and their relevance in processing visual information by humans. We draw the biological basis of our work from Burton et al (Burton et al., 1999) amongst others wherein the authors write – "PCA delivers information about the ways in which faces vary. It seems plausible that whatever representational scheme is used by humans in recognizing faces, the scheme captures the variance among its inputs". Therefore, the whole idea was to understand and then present a suitably chosen classifier those set of inputs that captures the aforesaid variance within the inputs. Hence, we thought it is worth delving in a design framework that uses PCA, PCA-LBP and PCA-HOG, and its fusion as a core system for "Face Recognition". The reasons for the selection of PCA alone, HOG, and LBP, as possible set for capturing the variance, are deliberated at length in Engineering Basis of Biological Process section.

Our experimental framework is quite comprehensive wherein we evaluated BIFR by following a model selection scheme and then quite extensively compared the results obtained from BIFR with state-of-the-art methods on benchmark datasets. We noticed that performance of BIFR is comparable and at times better by even up to 14.5% on earlier reported results.

Turk, M. & Pentland (Turk and Pentland, 1991) mentioned that there must also be some fast, low-level, two-dimensional image processing-based recognition mechanism that is more akin to human way of performing the task. Keeping in view the above point and other discussions as reported by Chowdhury, P. R. (Chowdhury, 2016), we deliberately choose to use MLP trained by scaled conjugate gradient method as a classifier for BIFR, and did not opt for other significantly improved ones like SVM. The reason for doing this is also motivated by the definition of "Thinking" given in Chowdhury, P. R. (Chowdhury, 2016), wherein we will need to re-construct brain's original neural network, to accomplish machine assisted thinking. We also argue that capturing relevant PCs in either image or feature space, are only a method to capture the key regularities in that space which happens in an unsupervised way. We in no way claim that there cannot be better operators than PCA, but on the contrary there may be. The whole objective of such a design and the comprehensive evaluation of BIFR were to keep the whole thing very simple and relevant to the context, and to observe the performance of such system regarding other methods.

Jo, J. & Bengio, Y (Jo and Bengio, 2017) pointed out that Deep Learning methods may have shallow understanding of the input space hence are susceptible to attacks that change the surface statistical regularities such as adversarial attacks and Fourier filters. Considering that we performed experiments on EYB dataset using a radial low pass Fourier filter of radius 40% of image size, accuracy of Xception decreased by 4% on average while accuracy of BIFR decreased by 0.7% on average, which further supports our claim of the proposed model being a brain inspired engineering equivalent of the functional mechanism for face recognition.

## 6.1. Comparison with convolution filters and V1 like features

We compare the results of AT&T dataset with results reported in Ritwik Kumar et al (2012) (Kumar et al., 2011) and others (Pinto et al., 2008). We specifically chose these methods as they are based on trainable convolution filters, and V1-like representation which is inspired by properties of cortical area V1. These methods are quite extensively used in modern days; therefore, we felt it prudent to compare performance of BIFR with the aforesaid methods.

Table 9 reports the comparison of BIFR on AT&T dataset with the results reported by Pinto et al (Pinto et al., 2008). In Table 10 we compare best recognition performance of BIFR with the results reported by Liu et al (Liu et al., 2018). In Nicolas Pinto et al (2008) (Pinto et al., 2008) and others (Liu et al., 2018) (for AT&T) the results are reported in a graphical format. Therefore, we approximate the recognition rates of these studies (Pinto et al., 2008),(Liu et al., 2018) in Table 9 and Table 10 below for comparisons.

TABLE 9

COMPARATIVE RESULTS SHOWING TESTING SET AVERAGE RECOGNITION ACCURACIES ON THE AT&T DATASET USING BIFR AND V1-LIKE FEATURES. * INDICATES THAT BEST OBTAINED RESULT IN 4 TRAIN AND 8 TRAIN CASES FOR AT&T ARE 98.00% AND 100% RESPECTIVELY. RESULTS ARE REPORTED IN PERCENTAGE. THE NAMES OF THE ALGORITHMS IN THE LEFT HAND COLUMN ARE THOSE REPORTED BY PINTO ET AL(Pinto et al., 2008).

| ALGORITHM | AT&T | |
|---|---|---|
| | 4 TRAIN | 8 TRAIN |
| Pixel space | 93.00 | 94.00 |
| Savvides et. al. 2007 | 96.00 | 97.00 |
| Noushath et al. 2006 | 95.00 | 98.50 |
| Ben et al. 2006 | - | - |
| Wang et.al 2007 | - | - |
| V1 –like(Pinto et al., 2008) | *98.00* | *`100* |
| **BIFR** | **94.70*** | **99.75*** |

TABLE 10

COMPARISON OF TESTING SET BEST RECOGNITION RESULTS ON THE ORL DATASETS USING BIFR AND VARIANTS OF PCA NET ALGORITHM. PERFORMANCE RESULTS ARE REPORTED IN PERCENTAGE.

| ALGORITHM | AT&T |
|---|---|
| | 5 TRAIN |
| PCA Net(Liu et al., 2018) | 98.00 |
| SPCA Net(Liu et al., 2018) | 97.50 |
| EPCA Net-offset(Liu et al., 2018) | *99.50* |
| EPCA Net-subsample(Liu et al., 2018) | 98.00 |
| **BIFR** | ***98.50*** |

## 7. Conclusion

This work examines the potential of a very simple framework of face recognition on large number of benchmark datasets. Since face regions in the brain are involved in general object recognition, we used the same model for object recognition tasks as well. We observe that the model is able to successfully distinguish face and non-face category and perform object recognition to a certain extent. Liu et al (Liu et al., 2018) beside others presented the importance of PCA in the domain of psychology at great details and its relevance in processing facial information by humans. This paper aptly demonstrates that (a) encapsulation of right domain knowledge always enhances the accuracy of systems designed using this knowledge, (b) features suitably delineated for a complex task like face recognition, when fed to a simple classifier, can even outperform methods like deep learning for that task, and (c) the best accuracy, once achieved, doesn't change significantly with number of PCs (Fig. 4(B)).

### Data availability

The data that support the findings of this study are available from the authors on reasonable request.

### Competing interests

The authors declare no competing interests.

### Funding

This research did not receive any specific grant from funding agencies in the public, commercial, or not-for-profit sectors.

### Author Contribution

**Pinaki Roy Chowdhury**: Conceptualization, Methodology, Formal Analysis, Investigation, Writing – Original Draft, Writing – Review & Editing, Supervision, Project administration. **Angad Singh Wadhwa**: Conceptualization, Methodology, Software, Validation, Formal Analysis, Investigation, Data curation, Writing – Original Draft, Visualization. **Nikhil Tyagi**: Conceptualization, Methodology, Software, Validation, Formal Analysis, Investigation, Data curation, Writing – Original Draft, Visualization